\documentclass[letterpaper, 10 pt, conference]{ieeeconf}  

\IEEEoverridecommandlockouts                              

\usepackage{graphicx}
\usepackage{amsmath} 

\title{\LARGE \bf
Uncovering Population PK Covariates from VAE-Generated Latent Spaces%
\thanks{© 20XX IEEE.  Personal use of this material is permitted.  Permission from IEEE must be obtained for all other uses, in any current or future media, including reprinting/republishing this material for advertising or promotional purposes, creating new collective works, for resale or redistribution to servers or lists, or reuse of any copyrighted component of this work in other works.}
}

\author{Diego Perazzolo$^{1,2}$, Chiara Castellani$^{1}$ and Enrico Grisan$^{2*}$
\thanks{$^{1}$Department of Cardiac, Thoracic, Vascular Sciences and Public Health, University of Padova, Via Giustiniani, 2 - 35128 Padova, Italy
        {\tt\small diego.perazzolo.1@phd.unipd.it}, {\tt\small chiara.castellani@unipd.it}}%
\thanks{$^{2}$School of Engineering, London South Bank University,
        103 Borough Road, London, SE1 0AA, UK
        {\tt\small enrico.grisan@lsbu.ac.uk}}%
\thanks{*Corresponding author}%
}

\begin{document}

\maketitle
\thispagestyle{empty}
\pagestyle{empty}

\begin{abstract}
Population pharmacokinetic (PopPK) modelling is a fundamental tool for understanding drug behaviour across diverse patient populations and enabling personalized dosing strategies to improve therapeutic outcomes. A key challenge in PopPK analysis lies in identifying and modelling covariates that influence drug absorption, as these relationships are often complex and nonlinear. Traditional methods may fail to capture hidden patterns within the data. In this study, we propose a data-driven, model-free framework that integrates Variational Autoencoders (VAEs) deep learning model and LASSO regression to uncover key covariates from simulated tacrolimus pharmacokinetic (PK) profiles. The VAE compresses high-dimensional PK signals into a structured latent space, achieving accurate reconstruction with a mean absolute percentage error (MAPE) of 2.26\%. LASSO regression is then applied to map patient-specific covariates to the latent space, enabling sparse feature selection through L1 regularization. This approach consistently identifies clinically relevant covariates for tacrolimus including SNP, age, albumin, and hemoglobin which are retained across the tested regularization strength levels, while effectively discarding non-informative features. The proposed VAE-LASSO methodology offers a scalable, interpretable, and fully data-driven solution for covariate selection, with promising applications in drug development and precision pharmacotherapy.
\newline

\indent \textit{Clinical relevance}— This study provides a fully data-driven framework for identifying key patient-specific factors influencing pharmacokinetics. By leveraging multiple techniques, the proposed methodology is highly adaptable, and applicable to multiple population pharmacokinetic studies, with the potential to improve therapeutic drug monitoring and patient treatment efficacy.  
\end{abstract}


\section{INTRODUCTION}
Pharmacokinetic (PK) aims to understand the absorption, distribution, metabolism, and excretion of drugs within the body.
Population pharmacokinetic (PopPK) modelling plays a crucial role in understanding drug behaviour across diverse patient populations, allowing for the optimization of individualized dosing regimens. Covariates such as age, weight, ethnicity, genetic factors and others often influence drug behaviour. With the growing emphasis on personalized treatment, reliable models are essential for identifying key covariates that drive variability in drug response \cite{schnider1998influence}.
Traditional PopPK modelling relies on the direct application of parametric regression techniques, which often struggle to capture the complex relationships between covariates and drug pharmacokinetics \cite{sherwin2012fundamentals} \cite{hutmacher2015covariate}. The clear identification of most relevant covariates is essential for improving therapeutic drug monitoring and ensuring optimal treatment. Recent advancements in machine learning have introduced deep generative models, such as Variational Autoencoders (VAEs), which can learn latent representations of PK profiles while preserving meaningful structure in high-dimensional data \cite{autoencoderDiederik} \cite{gomari2022variational}. In parallel, LASSO (Least Absolute Shrinkage and Selection) regression, a sparse linear model with L1 regularization \cite{ribbing2007lasso} \cite{muthukrishnan2016lasso}, has been widely adopted for feature selection in biomedical applications due to its ability to eliminate non-contributory variables and enhance model interpretability \cite{gamal2024novel}.
\newline
In this study, we introduce a data-driven VAE-LASSO framework for covariate selection in PopPK modelling of tacrolimus, without requiring prior knowledge of the underlying pharmacokinetic model and its parameters.
Tacrolimus is a commonly used immunosuppressant that plays a critical role in preventing graft rejection in patients who have undergone liver, kidney, or heart transplantation \cite{venkataramanan1995clinical}. It presents a unique challenge for PK modelling due to its narrow therapeutic index and high inter-individual variability \cite{chen2023population} \cite{nanga2019toward}.
The VAE is used to encode PK signals into a latent space representation, while LASSO regression maps patient-specific covariates to the latent space, enabling the direct identification of key predictors of PK profiles. We systematically analyzed the impact of different regularization strengths (\(\lambda\)) on covariate selection and assess the robustness of this approach in filtering out irrelevant ones. 
The findings of this study provide valuable insights into covariate selection by leveraging the latent representations learned through Variational Autoencoders. This approach enhances data-driven feature selection in PopPK modelling, with promising applications in precision dosing and personalized pharmacotherapy. Furthermore, we highlight the limitations of LASSO’s linear assumption and discuss future directions for integrating latent space modelling with non-linear techniques to improve the accuracy and robustness of pharmacokinetic analyses.

\begin{figure*}[thpb]
    \centering
    \includegraphics[width=0.8\textwidth]{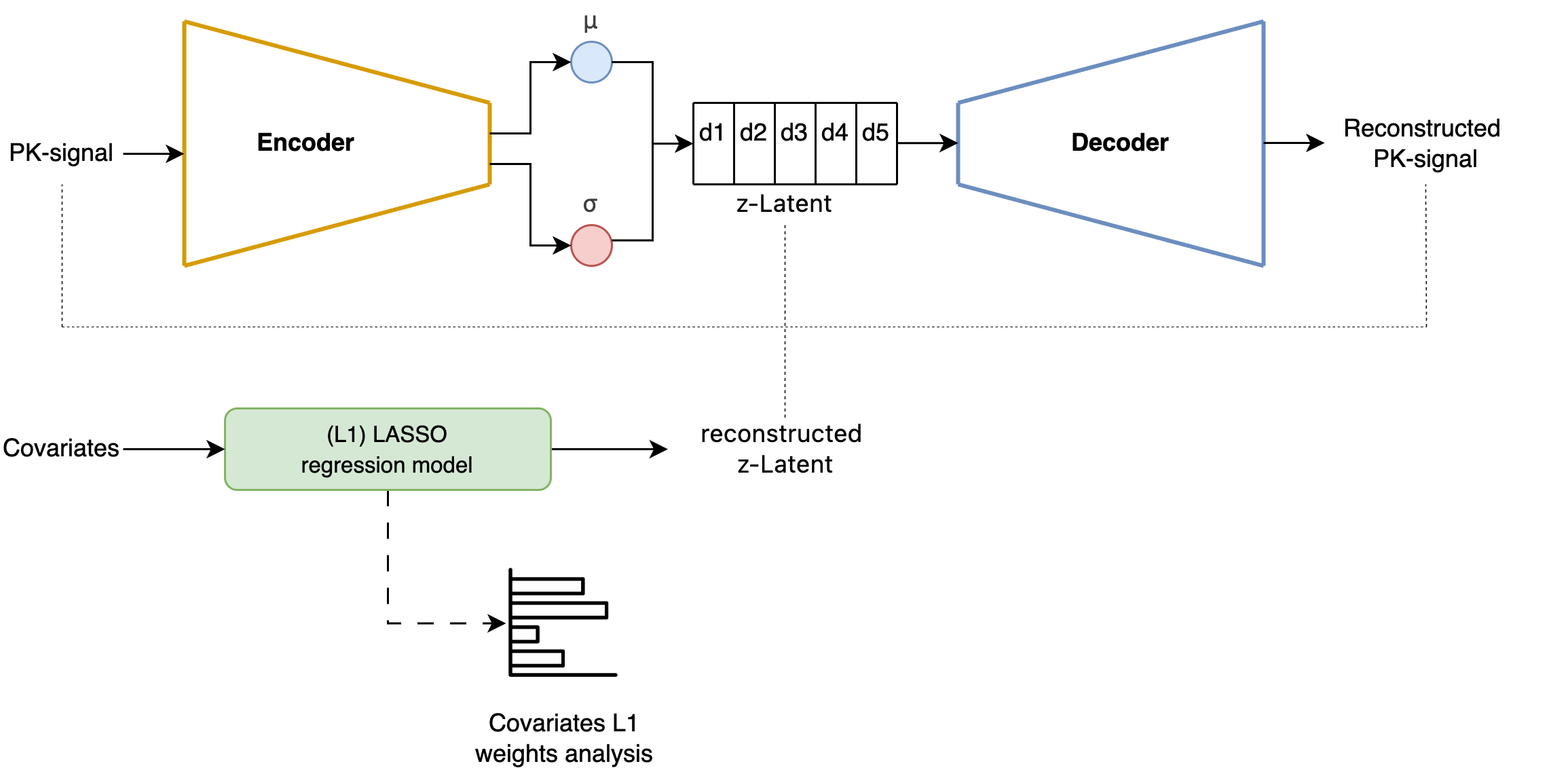}
    \caption{Graphical representation of the VAE-LASSO framework. 
    The Variational Autoencoder consists of an encoder (left) that compresses pharmacokinetic signals profiles into a latent representation parameterized by a mean (\(\mu\)) and standard deviation (\(\sigma\)). The latent variables are then sampled and passed to the decoder (right), which reconstructs the original PK profiles. A LASSO regression model (bottom) is trained to predict the latent representation using patient-specific covariates, allowing for covariate selection by shrinking irrelevant features to zero.}
    \label{vae_lasso_model}
\end{figure*}


\section{MATERIALS AND METHODS}

\subsection{Data Generation}
\subsubsection{Pharmacokinetics Modelling of Tacrolimus}
The tacrolimus concentration at each time step, have been estimated using a one-compartment pharmacokinetic model depicted from (\ref{one_compartment_elimination_equation}). The generation of synthetic PK profiles was carried out in accordance with the settings outlined in the study by Chen et al. \cite{chen2023population}. We generate Fast-Elimination PK profiles, with fixed dose $D$ of 300 (mg) and elimination rate ($K_{e}$) calculated as $K_{e} = CL/V$, where $V$ is the volume of distribution. The absorption rate constant $k_{a}$ was fixed to 0.502 $(h^{-1})$ and the absorption time lag ($t_{lag}$) was fixed to 0.346 $(h)$.

\begin{equation}
\begin{split}
    C(t) &= \frac{D}{V} \times \frac{k_a}{k_a - k_e} \times \\
         &\quad \left( e^{-k_e (t - t_D - t_{\text{lag}})} 
         - e^{-k_a (t - t_D - t_{\text{lag}})} \right)
\end{split}
\label{one_compartment_elimination_equation}
\end{equation}


\vspace{5mm}
The equation (\ref{clearence_equation}) is used to measure clearance $CL$ for subject $i$ at time $j$. In the equation, $X_{1}$ corresponds to \textit{SNP} (Single Nucleotide Polymorphism) of the CYP3A5 genotypes. Following the encoding strategy adopted by Chen et al., the three genotypic variants (expressor, intermediate expressor, and non-expressor) are numerically coded as 1, 2, and 3, respectively. This numerical representation allows the inclusion of genetic polymorphism as a continuous covariate in the clearance model. $X_{2}$ refers to \textit{age} as the subject age in year, $X_{3}$ is the albumin (\textit{alb}) level in g/dL and $X_{4}$ the blood haemoglobin  (\textit{hgb}) concentration in g/dL of the subject i at time j. Volume has been estimated using (\ref{volume_equation}). 

\begin{equation}
CL_{ij} = \theta_1 \times X_1^{\theta_2} \times X_2^{\theta_3} \times X_3^{\theta_4} \times X_4^{\theta_5} \times \textit{exp}^{[nCL]}
\label{clearence_equation}
\end{equation}

\begin{equation}
V_{ij} = \theta_6 \times \textit{exp}^{[nV]}
\label{volume_equation}
\end{equation}

where:
\begin{itemize}
    \item $\theta_1 = 26.2$\\
    \item $X_1 = (SNP_{ij})$ and $\theta_2 = 0.71$ \\ 
    \item $X_2 = (\frac{age_{ij}}{47})$ and $\theta_3 = -0.26$\\
    \item $X_3 = (\frac{alb_{ij}}{4.1})$ and $\theta_4 = 0.35$ \\
    \item $X_4 = (\frac{hgb}{125})$ and $\theta_5 = -0.29$\\
    \item $\theta_6 = 3726$\\
\end{itemize}

\subsubsection{Covariates}
The covariates used to generate the PK profiles include age, sex (male, female), weight, haemoglobin, albumin, CYP3A5 SNP (with three variants coded as 1,2,3), and ethnicity (race), categorized into Caucasian-American, African-American, Hispanic, Asian and other. To further assess the robustness and discriminative capacity of the covariate selection mechanism, we additionally introduced two random covariates (extra\_1 and extra\_2), sampled from a uniform distribution in the range between 0 and 1. These synthetic variables were intentionally designed to be completely uninformative with respect to the PK profiles. Their inclusion serves as a negative control to verify whether the system is capable of correctly disregarding irrelevant covariates. Categorical covariates were sampled using a uniform random distribution to ensure equal probability among categories. Continuous variables were collected, sampling from a Gaussian normal distribution, with each covariate assigned a specific mean (\(\mu\))  and standard deviation (\(\sigma\)) (listed below). 
\newline
\begin{itemize}
    \item age: $\mu = 45.9$, $\sigma = 12.7$ 
    \item weight: $\mu = 82.9$, $\sigma = 20.8$ 
    \item haemoglobin: $\mu = 12.5$, $\sigma = 2.1$ 
    \item albumin:$\mu = 4.1$, $\sigma = 0.4$\\
\end{itemize}

\subsubsection{Synthetic Data Generation}
A total of 10,000 PK profiles, representing tacrolimus concentrations (mg/L) over a 48-hour period following dose administration, were created. Each with a specific set of covariates. In equation \ref{clearence_equation} and \ref{volume_equation}, $\textit{exp}^{nCL}$ and $\textit{exp}^{nV}$ represent the random effects of between-subject variability. These random effects were sampled from a normal distribution with a mean of 0 and standard deviations of 0.408 and 0.653, respectively. These pharmacokinetic signals were used to train the VAE, in order to construct a latent representation of all of them. 
A separate test set of 2,000 PK profiles was generated following the same procedure to assert the VAE model's PK profile reconstruction ability and generalization.

\subsection{Models Architecture}
The proposed framework consists of a Variational Autoencoder trained to learn a latent representation of simulated PK profiles, followed by a Least Absolute Shrinkage and Selection Operator (LASSO) regression model that performs covariate selection, identifying those influencing the latent space reconstruction. The representation of the entire framework architecture can be seen in Figure \ref{vae_lasso_model}. 
\newline

\subsubsection{PK Variational Autoencoder (VAE)}
Unlike a standard autoencoder, which compresses input data into a deterministic latent space, a VAE encodes the input into a probabilistic latent distribution. This probabilistic nature addresses the issue of non-regularized latent space of autoencoders and provides the generative capability to the entire space. The encoder maps PK profiles into a lower-dimensional latent space characterized by a mean ($\mu$) and a standard deviation ($\sigma$), which parameterize a Gaussian distribution from which latent variables are sampled. The decoder reconstructs PK profiles from these latent variables, optimizing a combined loss function consisting of the reconstruction error and the Kullback-Leibler (KL) divergence, which regularizes the latent space distribution \cite{autoencoderDiederik}. 
\newline

\subsubsection{Latent Space Regression}
The LASSO regression model determines the influence on each covariate in the latent space reconstruction by adding a regularization penalty term the L1 regularization. The penalty term forces regression coefficients to become exactly zero, performing data-driven feature selection. \cite{muthukrishnan2016lasso}. 

\begin{equation}
\hat{\beta} = \arg\min_{\beta} \left\{ \sum_{i=1}^{n} \left( y_i - \sum_{j=1}^{p} X_{ij} \beta_j \right)^2 + \lambda \sum_{j=1}^{p} |\beta_j| \right\}
\label{lasso_equation}
\end{equation}

Equation (\ref{lasso_equation}) depicts the classical ordinary least squares (OLS) function used to estimate the coefficient vector $\hat{\beta}$ that represents the weights assigned to each predictor (i.e. each covariate). $\lambda\sum_{j=1}^{p}|\beta_{j}|$ is the L1 regularization penalty term, that controls the sparsity of the model. $\lambda$ is the crucial parameter that 
determine the strength of the penalty term.

\subsection{Experimental Procedure}

\subsubsection{Covariate Preprocessing}
Prior to model training, categorical covariates were processed using one-hot encoding, while continuous covariates were normalized using min-max scaling to ensure consistency across features. 
\newline
\subsubsection{VAE Model Training}
\begin{figure}[htbp]
    \centering
    \includegraphics[width=\columnwidth]{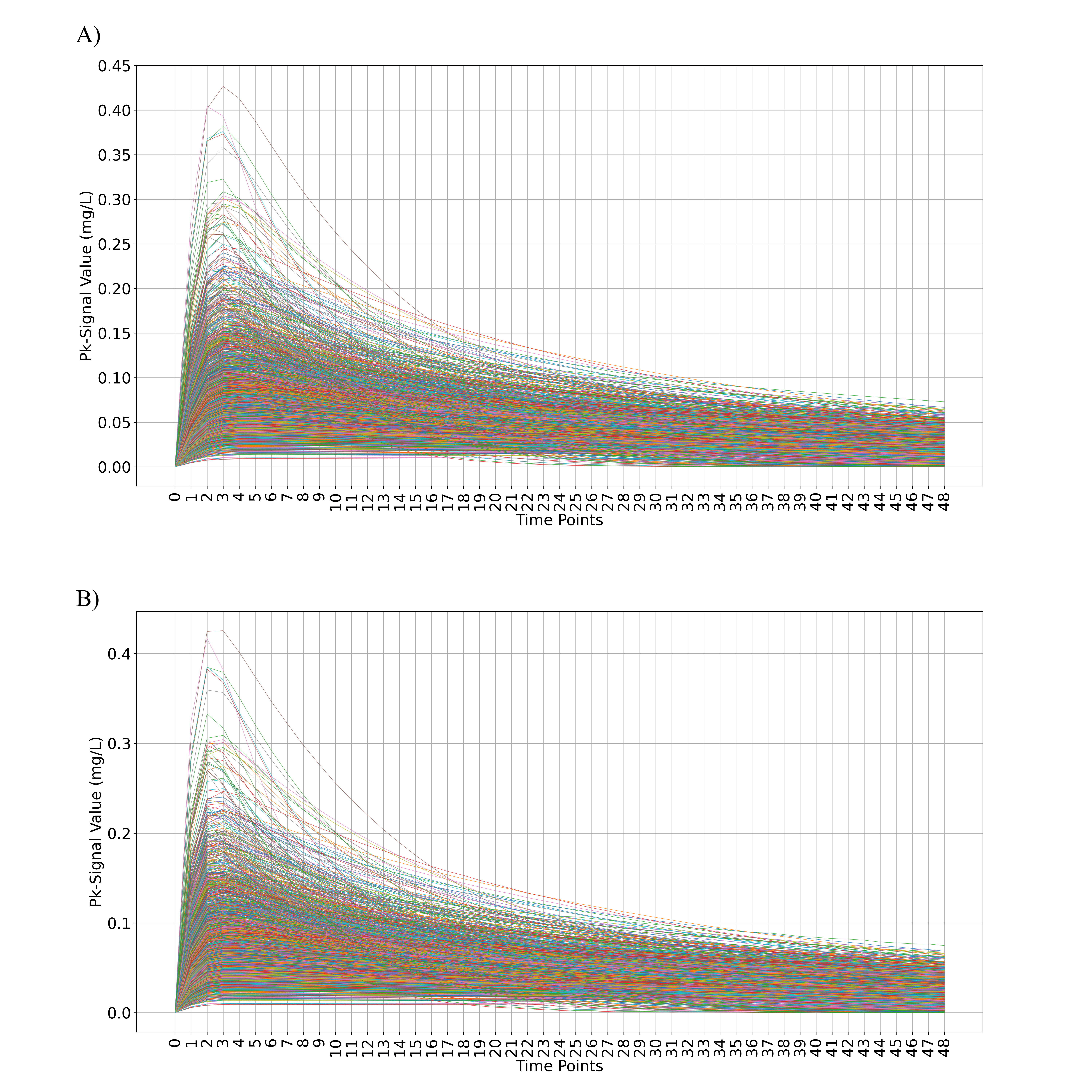}
    \caption{(A) Original tacrolimus PK profiles generated from the simulated dataset. (B) Reconstructed PK profiles obtained from the Variational Autoencoder (VAE). The y-axis represents tacrolimus concentration in mg/L over time. The optimal reconstruction that closely follow the original data, demonstrates  the VAE’s ability to capture the underlying PK dynamics.}
    \label{VAE_pk_test_reconstruction}
\end{figure}

The VAE model was trained using the 10,000 PK profiles as input, to obtain a representative latent space. The training process was optimized using mean absolute error (MAE) as reconstruction loss to minimize differences between input and output PK signals and a KL-divergence term to enforce smoothness in the latent space distribution.
We evaluated VAE architecture on the generated test set of 2,000 tacrolimus PK profiles. The model achieve an optimal mean absolute percentage error (MAPE) of $2.26\%$ (measured metrics are collected in Table \ref{table_reconstruction_error}) in PK-profiles reconstruction. Metrics have been measured using Python’s scikit-learn library. Figure \ref{VAE_pk_test_reconstruction} shows the comparison of the test set (panel A) and the corresponding reconstructed profiles generated by the VAE (panel B). Each line represents one of the 2,000 individual PK profiles.

\begin{table}[h]
\caption{PK profiles reconstruction error from VAE}
\label{table_reconstruction_error}
\begin{center}
\begin{tabular}{|c||c|}
\hline
Metric & VAE model\\
\hline
MAE (Mean Absolute Error) & 0.0009 $(mg/L)$\\
\hline
MAPE (Mean Absolute Percentage Error) & 2.26 \%\\
\hline
\end{tabular}
\end{center}
\end{table}

\subsubsection{LASSO Regression model}
The LASSO regression model was developed using Python’s scikit-learn library \cite{scikit-learn}. It was applied to establish a direct mapping between patient-specific covariates and the latent space representations of the 10,000 PK-tacrolimus profiles used as training set for the VAE. To evaluate the impact of regularization strength on covariate selection, different values of the $\lambda$ parameter were tested. Lower $\lambda$ values allowed for a larger number of covariates to be retained, whereas higher $\lambda$ values forced greater sparsity, removing less relevant features. For each tested \(\lambda\) value, the weights assigned to each covariate were collected and analyzed. We systematically explored a suitable range of regularization strengths and conducted an in-depth analysis of covariate selection for the following \(\lambda\) values: \(0.0001, 0.002, 0.005, 0.008, 0.01, 0.1, 1.0\).



\section{RESULTS}
\begin{figure}[t]
    \centering
    \includegraphics[width=\columnwidth, height=0.4\textheight]{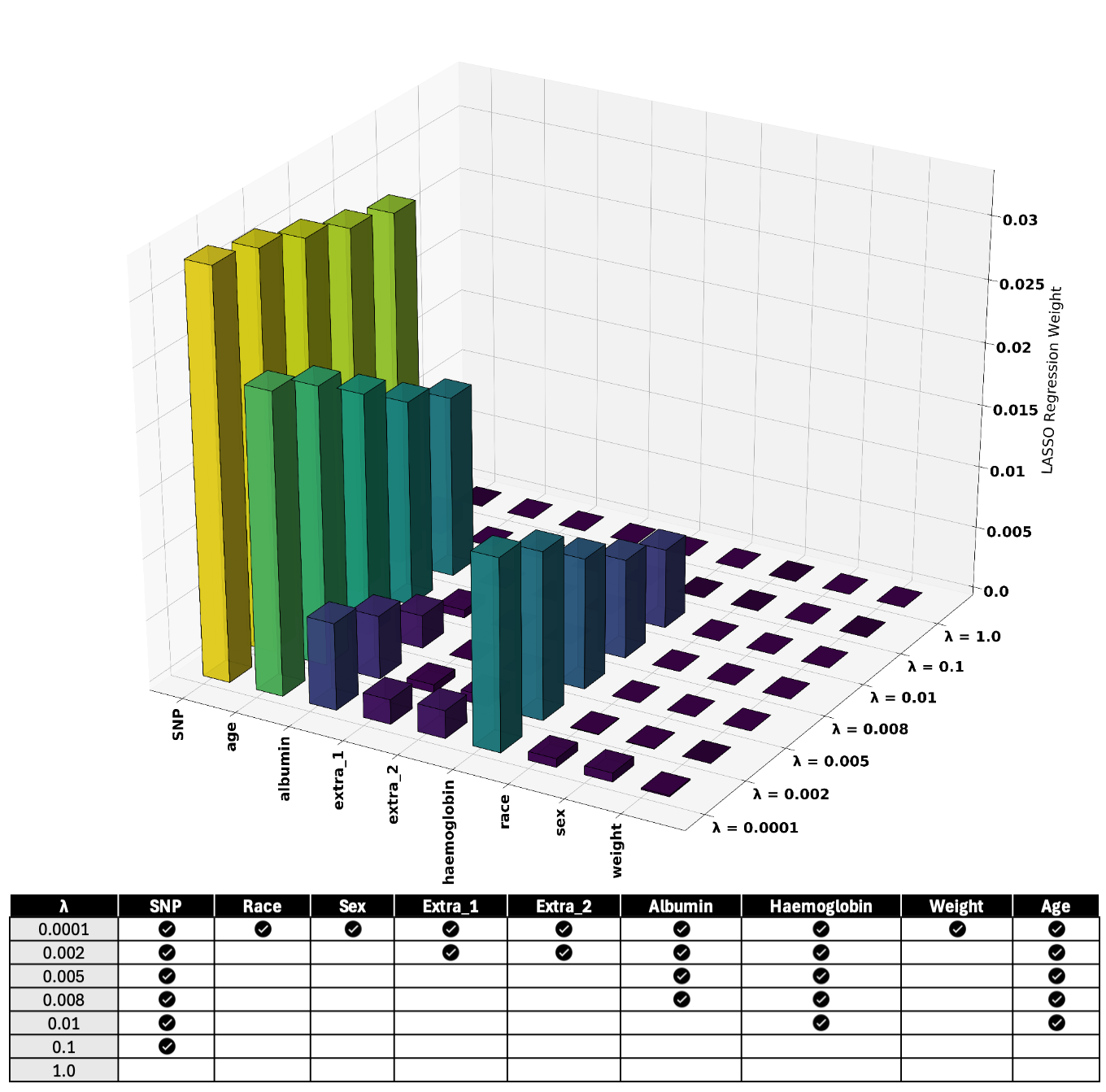}
    \caption{3D Visualization of LASSO Regression weights wcross covariates and lambda values. The height and color intensity of the bars represent the importance of each covariate at different levels of regularization \(\lambda\). Brighter and taller bars indicate covariates with stronger influence (i.e., higher regression weights), while darker, shorter bars correspond to covariates with little or no contribution.The x-axis lists the covariates, y-axis the different values of the regularization parameter \(\lambda\) and the z-axis the LASSO regression weights. The table denote with markers the covariates retained at each regularization level.}
    \label{3D_representation_cov_selection}
\end{figure}

Our results demonstrate that the LASSO regression model effectively identifies key covariates in equation \ref{clearence_equation} that characterize tacrolimus pharmacokinetic profiles by reconstructing the latent space generated by the VAE. Figure \ref{3D_representation_cov_selection} illustrates the covariates selected by the model and the effect of varying the regularization parameter \(\lambda\). As \(\lambda\) increases, the model progressively eliminates weaker covariates and the two randomly generated noise variables (extra\_1 and extra\_2), confirming the robustness of LASSO in filtering out non-contributory features.
The SNP, age, hemoglobin, and albumin covariates are consistently retained across different \(\lambda\) values, emphasizing their fundamental role in tacrolimus pharmacokinetics. The optimal \(\lambda\) value is determined by balancing model interpretability with feature selection efficiency. Lower \(\lambda\) values allow more covariates to contribute, potentially capturing minor effects, whereas higher \(\lambda\) values enforce greater sparsity, retaining only the most influential predictors.
To further assess the effectiveness of LASSO-based covariate selection, we utilized the reconstructed latent space to decode PK profiles via the VAE decoder. However, the reconstructed PK profiles exhibited insufficient accuracy, primarily due to the linear nature of LASSO regression. Since pharmacokinetic signals are governed by complex, non-linear dynamics, the simplified linear mapping imposed by LASSO does not allow the complete correct reconstruction of the PK profiles. Despite this limitation, our findings indicate that the LASSO-reconstructed latent space provides an efficient structured and interpretable representation that enables efficient covariate selection.

\section{DISCUSSIONS AND CONCLUSIONS}
Our findings demonstrate that the proposed VAE-LASSO framework effectively integrates latent space modelling with sparse regression techniques for covariate selection in population pharmacokinetics (PopPK). Unlike traditional PopPK approaches, which rely heavily on parametric regression models with predefined structural assumptions, our method is model-free and data-driven, requiring no prior knowledge of the underlying pharmacokinetic equations or covariate interactions. This distinction is crucial, as traditional PopPK models often rely on stepwise covariate selection procedures such as forward selection or backward elimination which, despite their simplicity, are prone to overfitting and may fail to capture complex, nonlinear relationships in high-dimensional data. In contrast, our approach leverages a Variational Autoencoder to learn a compact and structured latent representation directly from the pharmacokinetic profiles, enabling the discovery of hidden patterns without requiring prior assumptions about the underlying model structure. Despite its effectiveness in feature selection, the linear nature of LASSO regression limits its ability to fully reconstruct the latent space and, consequently, the pharmacokinetic profiles. This limitation arises from the non-linearity inherent in PK systems, which LASSO fails to capture. Nevertheless, the model provides a structured and interpretable latent representation that simplifies covariate selection, making it a valuable tool for clinical decision-making. It also highlights the potential of latent space representations in uncovering underlying pharmacokinetic patterns, facilitating a data-driven approach to identifying key patient-specific factors influencing drug metabolism. 
By leveraging the outcome of this study, future works will aim to further explore the integration of latent space modelling with other multiple non-linear based techniques and improving the interpretability of latent space representations for enhance therapeutic decision-making. Overall, this study demonstrates the potential of integrating deep learning and sparse regression to enhance covariate selection in pharmacokinetic modelling, paving the way for more robust and interpretable machine learning approaches in clinical pharmacology.

\addtolength{\textheight}{-5cm}   







\bibliographystyle{IEEEtran}  
\bibliography{biblio_IEEE}  

\end{document}